\documentclass[conference]{IEEEtran}
\IEEEoverridecommandlockouts
\usepackage[]{subfigure,epsfig,amsthm,enumerate, array, latexsym,paralist,bm,bbm,color,wrapfig,multirow}
\usepackage{algpseudocode}
\usepackage{xcolor}
\usepackage{cite}
\usepackage{amsmath,amssymb,amsfonts}
\usepackage{graphicx}
\usepackage{textcomp}
\usepackage{hyperref}
\usepackage{datetime} 
\newcommand{\Pro}{\mathbb{P}} 

\newcommand{\R}{\mathbb{R}} 
\newcommand{\W}{\mathbb{W}} 
\newcommand{\D}{\mathcal{D}} 
\def\X{\mathbb{X}}

\def\0{\mathbf{0}}

\def\beqn{\begin{equation}}
\def\eeqn{\end{equation}}
\def\bean{\begin{eqnarray}}
\def\eean{\end{eqnarray}}
\def\bea{\begin{eqnarray*}}
\def\eea{\end{eqnarray*}}

\newcommand{\rhighlight}[1]{%
	\colorbox{blue!30}{$\displaystyle#1$}}
\newcommand{\bhighlight}[1]{%
	\colorbox{green!20}{$\displaystyle#1$}}
\newcommand{\brhighlight}[1]{%
	\colorbox{yellow!40}{$\displaystyle#1$}}

\def\BibTeX{{\rm B\kern-.05em{\sc i\kern-.025em b}\kern-.08em
    T\kern-.1667em\lower.7ex\hbox{E}\kern-.125emX}}
\begin{document}

\title{Bayesian Topological Learning for Brain State Classification }
\author{Farzana Nasrin, Christopher Oballe, David L. Boothe, and Vasileios Maroulas
	\thanks{ F. Nasrin. C. Oballe and V. Maroulas are with University of Tennessee, Department of Mathematics, Knoxville, TN, USA (email: fnasrin@utk.edu). D. L. Boothe is with US Army Research Laboratory, Aberdeen Proving Ground, MD 21005, USA. The  authors thank Piotr J. Franaszczuk from US Army Research Laboratory for his insightful comments and the software that generates dataset considered in this work.}
\thanks{FN's research was supported by the ARO  W911NF-17-1-0313.
	CO's research was sponsored by the Army Research Laboratory (ARL) and was accomplished under Cooperative Agreement Number W911NF-19-2-0302.
	DB's work was supported by the Combat Capabilities and Development Command US ARL.
	VM's work was partially supported by the ARO  W911NF-17-1-0313, NSF MCB-1715794 and DMS-1821241, Thor Industries/ARL  W911NF-17-2-0141, and ARL and was accomplished under Cooperative Agreement Number W911NF-19-2-0328. The views and conclusions contained in this document are those of the authors and should not be interpreted as representing the official policies, either expressed or implied, of the Army Research Laboratory or the U.S. Government. The U.S. Government is authorized to reproduce and distribute reprints for Government purposes notwithstanding any copyright notation herein.}}


\maketitle
\begin{abstract}
Investigation of human brain states through electroencephalograph (EEG) signals is a crucial step in human-machine communications. However, classifying and analyzing EEG signals are challenging due to their noisy, nonlinear and nonstationary nature.
 Current methodologies for analyzing these signals often fall short because they have several regularity assumptions baked in. This work provides an effective, flexible and noise-resilient  scheme to analyze EEG by extracting pertinent information while abiding by the 3N (noisy, nonlinear and nonstationary) nature of data. We implement a topological tool, namely persistent homology, that tracks the evolution of topological features over time intervals and incorporates individual's expectations as prior knowledge by means of a Bayesian framework to compute posterior distributions. Relying on these posterior distributions, we apply  Bayes factor classification to noisy EEG measurements. The performance of this Bayesian classification scheme is then compared with other existing methods for EEG signals.
\end{abstract}


\section{Introduction}
The emergence of computational intelligence has led us to an era of 
excellent communication between users and systems. These human-computer communications do not require any external device or muscle intervention and enable computers to be deliberately controlled via the monitoring of brain state signals. In order to potentially improve human-machine interactions, it is crucial to analyze and interpret physiological measurements effectively to assess individual's states \cite{Ilyas2016, Padfield2019}. 
Brain signals can encode one's expectations as a form of \emph{prior beliefs}, which have an influence on behavior in times of uncertainty \cite{Sohn2019}. A Bayesian approach that integrates prior knowledge of an individual’s innate brain activity with newly measured data may improve individual's state detection, which can aide to characterize and control one’s actions. 

EEG signals are 3N--nonstationary, nonlinear and noisy \cite{Klonowski2007}. In particular, they are obscured by various forms of noise, 
are nonlinear due to the complexity of underlying interaction in the nervous system \cite{Stam2005,Klonowski2007,Klonowski2004} and are nonstationary due to the involvement of different time scales in brain activity \cite{Indic1999}. The 3N nature of EEG signals requires methods that can encode individual's brain history and draw statistical inferences for these signals.

In this paper, we develop a Bayesian classification scheme relying on the posterior distributions of persistence diagrams, which are pertinent topological descriptors. 
 Persistent homology is a widely used tool for topological data analysis (TDA) that captures topological features at multiple scales and produces  summary representations, called persistence diagrams, that encode the lifespan of the topological features in the data. Persistent homology has proved to be promising in the field of data sciences yielding astounding results in a variety of applications in variety of applications \cite{Patrangenaru2018, Guo2018,Sizemore2018,Babichev2017,Biscio2019,Maroulas2019,Sgouralis2017,Marouls2015,Mike2016,Marchese2018,Marchese2016,Lee2017,Ichinomiya2017,Kimura2018,Maroulas2019a,Humphreys2019}.
Indeed, physiological signals' features are defined by the topological changes of the signals across time. 
Engaging TDA in the study of physiological signals is recently emerging. The authors of \cite{Wang2018} measure the topology of EEG data with persistence landscapes to detect differences in the EEG signals during epilepsy attacks versus those produced during healthy brain activity.
 However, this method does not investigate the distribution of the diagrams themselves and suffers from a loss of pertinent information. 
Several other studies implement traditional machine learning based on feature extractions \cite {Dindin2019,Wang2017,Piangerelli2018}. As selection of an appropriate feature is crucial, these methods rely on summaries of persistence diagrams, which already summarize the underlying data themselves. 
We develop a Bayesian learning approach that can be applied \emph{directly} on the space of persistence diagrams. However, this learning scheme depends on the estimation of posterior probabilities, which is not straightforward due to the unusual multiset structure of persistence diagrams.


To establish a Bayesian framework, we need to define the prior uncertainty and likelihood through probability distributions of persistence diagrams. By viewing persistence diagrams as finite set samples, the authors of \cite{Maroulas2019}  propose a nonparametric estimation of the probability density function of random persistence diagrams. They also show that the probability density function can successfully detect the underlying dynamics of EEG signals and compare it with other pre-existing TDA methods.  A prior distribution of persistence diagrams can be obtained through this density function. However, computing posteriors entirely through the random set analog of Bayes' rule may have exponential computational complexity \cite{Goodman1997}.
To address this, we model random persistence diagrams as 
 point processes. In particular,  we utilize Poisson point processes which can be entirely characterized by their intensity. 

 
We commence the Bayesian framework by modeling random persistence diagrams generated from a Poisson point process with a given intensity, which captures the prior uncertainty. In the case of brain state detection, we can incorporate individual's expectations about the statistical regularities in the environment as prior knowledge \cite{Sohn2019}. Alternatively, we can choose an uninformative prior intensity when no expert opinion or information about individual's expectations is available. We construct the likelihood component of our framework by utilizing the theory of marked point processes. Indeed, we employ the topological summaries of signals in place of the actual signals. This 
proves to be useful for a range of physiological signal analysis \cite{Dindin2019,Wang2017,Piangerelli2018,Wang2018,Altindis2018,Campbell2019,Ilyas2016}. The application considered in this paper is the classification of EEG signals, which allows us to predict individual's brain states and advances human-computer communication techniques. Through these topological summaries, we adopt a substitution likelihood technique \cite{Jeffreys1961} rather than considering the full likelihood of the entire signal data. 

 Next, we develop a Bayesian learning method by relying on the  posterior obtained from the Bayesian framework. 
 This method is remarkably flexible as it abides by the 3N nature of the signals and is extremely powerful as it incorporates individual's expectations or domain experts' knowledge as prior beliefs. Furthermore, the Bayes factor provides a measure of confidence that in turn dictates whether further investigation is feasible.
 Our model enjoys a closed form of the posterior distribution through a conjugate family of priors, e.g., the Gaussian mixtures. Hence the prior-to-posterior updates yield posterior distributions of the same family. We present a detailed example of our closed form implementation on simulated EEG signals to demonstrate computational tractability and showcase applicability in classification through Bayes factor estimation. Furthermore, we present a detailed
 comparison with other TDA and non-TDA based learning methods.
 
%
%

This paper is organized as follows. 
Section \ref{sec:prelm} provides a brief overview of persistence diagrams and Poisson point processes. We establish the Bayesian framework for persistence diagrams in Section \ref{sec:main}. We then develop our Bayesian learning
method in Section \ref{sec:class}, which is used to quantify the classification outcome. Section \ref{sec:gm_post} introduces a closed form to the posterior intensity utilizing Gaussian mixture models. To assess the capability of our algorithm, we investigate its performance in  classifying EEG signals and 
provide comparisons with several other existing methods in Section \ref{sec:app}.  
Finally, we end with the conclusion in Section \ref{sec: conclusion}.

\section{Background \label{sec:prelm}}

We commence by discussing the background essential for building our Bayesian model. In Subsection \ref{sec:sublevel}, we start with the formation of persistence diagrams (PDs) by implementing sublevel set filtrations. In order to model the uncertainty present in these persistence diagrams, we consider them as point processes and pertinent definitions from point processes (PPs) are given in Subsection \ref{sec:ppp} .
	
\subsection{Persistent Homology for Noisy Signals \label{sec:sublevel}}
Persistent homology is a tool from TDA that provides a robust way to model the topology of real datasets by tracking the evolution of homological features and summarizing these in persistence diagrams. 
 Several methods exist to generate persistence diagrams such as Vietoris Rips or $\check{\text{C}}$ech filtrations  \cite{Edelsbrunner2010}, but such techniques require the transformation of a signal to an appropriate point cloud using Takens's delay embedding theorem. To circumvent this transformation to point clouds, we employ the sublevel set filtration method, which summarizes the shape of signals \emph{directly} in a PD by employing local critical points as tersely outlined next.  

\begin{figure}[b]
	\centering
	\subfigure[]{\includegraphics[width=2.3in,height=1.2in]{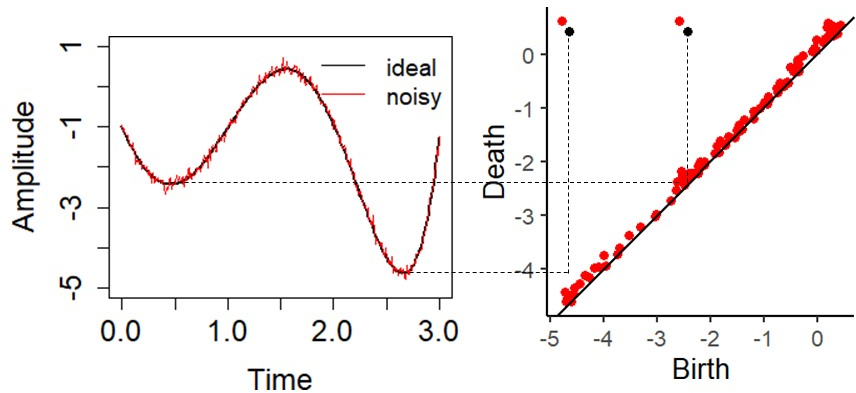}}
	\subfigure[]{\includegraphics[width=1.1in,height=1.1in]{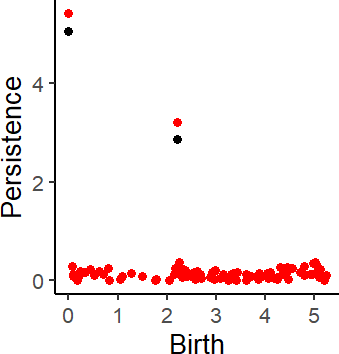}}
	\caption{(a) is illustrating the conversion of signals to the corresponding PDs using sublevel set filtrations. A smooth signal and a noisy version of it are presented in black and red respectively. For persistence diagrams, we make consistent color choices to instantiate the robustness of persistent homology to noise. (b) is the tilted representation of the PD in (a)}
	\label{fig:noise_robustness}	
	\vspace{-0.1in}
\end{figure}

Consider a signal as a bounded and continuous function of time $f(t)$ (Fig. \ref{fig:noise_robustness} (a)). The sublevel set filtration tracks the evolution of connected components in sets $f^{-1}((-\infty,r])$, as $r$ increases.  The central idea is that as $r$ increases the connectivity of the set $f^{-1}((-\infty,r])$ remains unchanged except when it passes through a critical point. For a given connected component, we record the value of $r$ at which is born (when $r$ reaches a local minimum), call it ${b}$, and the value at which  disappears (when $r$ reaches a local maximum), call it ${d}$, by merging with a pre-existing connected component. That is to say, whenever two connected components merge, the one born later disappears while the one born earlier persists by the elder rule \cite{Edelsbrunner2010}. Once we reach the value $\max f(t)$ in the filtration, all the sublevel sets have merged into a single connected component, and we terminate the procedure. For every connected component that arises in the filtration, we plot the points $({b},{d})$ in $\mathbb{R}^{2}$ and call the resulting collection a persistence diagram (Fig. \ref{fig:noise_robustness} (b)). 
To facilitate computation and preserve the geometric information, we apply the linear transformation $(b,p)=T(b,d) = (b-\min(b),d-b)$ to each point in our persistence diagrams. We refer to the resulting coordinates as birth and persistence, respectively, in  $\W := \{(b,p) \in \R^{2} | \,\, b,p \geq 0\}$ and call this transformed persistence diagram a tilted representation (Fig. \ref{fig:noise_robustness} (c)). Hereafter whenever we refer to persistence diagrams, we imply their tilted representation. 


\subsection{Poisson Point Processes \label{sec:ppp}}


 One samples from a finite point process $\mathcal{P}$ on a Polish space $\X$ by generating a random number $N$ according to a cardinality distribution and then for $N=n$ spatially distribute $\mathbf{x}=(x_1, \cdots,x_n) \in \X$ according to a probability distribution.  
  In other words, a finite point process is characterized by a probability mass function (pmf) of the cardinality and a joint probability density function (pdf) of the elements for a given cardinality. We model random persistence diagrams as Poisson point processes (PPPs), hence as points $\mathbf{(b,p)}=\mathbf{x} \in \W$.  The defining feature of these point processes is that they are solely characterized by a single parameter known as the intensity. The intensity $\lambda(x)$ of a given $x \in \R^d$ is the density of the expected number of points per unit volume at $x$. Indeed, the intensity serves as an analog of the first order moment of a random variable. The intensity in a Poisson point process accounts for the joint pdf of elements, and the cardinality is Poisson with mean $\mu = \int \lambda(x)$.
 
 Considering persistence diagrams as modeled by such processes, a link is needed between the prior and the data/likelihood to conduct Bayesian analysis. 
 The \emph{marked point process} provides this connection. 
 Effectively, a marked point process is a special case of bivariate point process where one PP $\Psi_M$ in the Polish space $\W_M$ (containing the marks) is determined given knowledge of the PP $\Psi$ in the Polish space $\W$. 
 	A marked Poisson point process $\Psi_{M}$ is a finite PP on $\mathbb{W} \times \mathbb{W}_M$ such that: 
 	(i) $\Psi=\left(\left\{p_{n}\right\},\left\{\Pro_{n}(\bullet)\right\}\right)$ is a PPP on $\mathbb{W}$, and 
 	(ii) for a realization  $(\mathbf{x},\mathbf{m}) \in \mathbb{W} \times \mathbb{W}_M$, the marks $m_i \in \mathbf{m}$  of each $x_i \in \mathbf{x}$ are drawn independently from a given stochastic kernel $\ell(\bullet|x_i)$.

%
%

\section{The Bayesian Model \label{sec:main}}
%
According to Bayes' theorem, the posterior is proportional to the product of a likelihood function and a prior. To investigate Bayesian framework for persistence diagrams, we need to compute the conditional distribution $p({D}_X \vert {D}_Y)$ by establishing the proposed Bayesian formula for persistence diagrams, 
$p({D}_X \vert {D}_Y)  \propto  \mathcal{L}({D}_Y | {D}_X) p({D}_X),$
where the likelihood $\mathcal{L}({D}_Y | {D}_X)$ and the prior $p({D}_X)$ need to be defined and computed for random persistence diagrams. We employ a likelihood model for the persistence diagrams generated from the signals which is analogous to the idea  of substitution likelihood \cite{Jeffreys1961}.
Next, we develop the prior and likelihood on the space of persistence diagrams.

\begin{figure}[t]
	\centering
	\subfigure[]{\includegraphics[width=1in,height=1.5in]{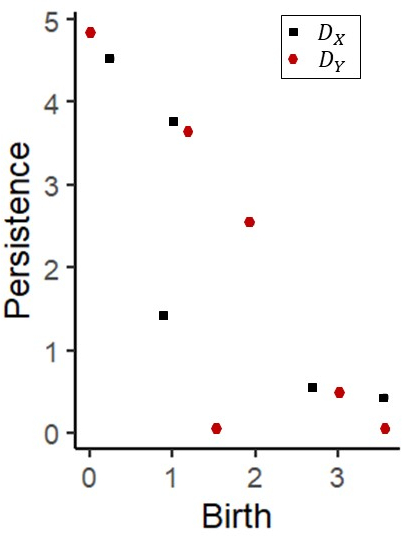}}\hspace{0.1in}
	\subfigure[]{\includegraphics[width=1in,height=1.5in]{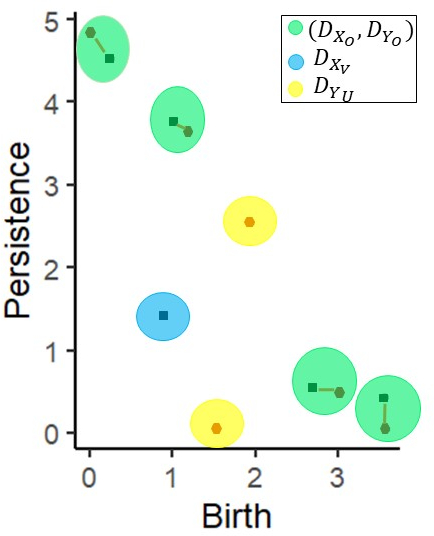}} \hspace{0.1in}
	\subfigure[]{\includegraphics[width=1in,height=1.5in]{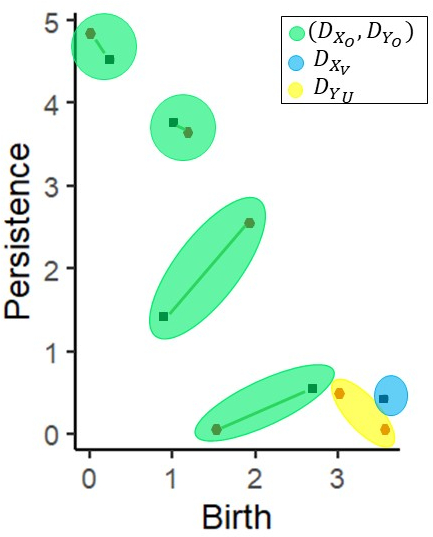}}\hspace{0.1in}
	\vspace{-0.1in}
	\caption{(a) is a sample $D_X$ from prior Poisson PP $\D_X$ and an observed persistence diagram $D_Y$. (b) and (c) are the decomposition of $D_X$ into $D_{X_O}$ \& $D_{X_V}$ and   $D_Y$ into $D_{Y_O}$ \& $D_{Y_U}$.}
	\label{fig:BayesTDA}	
		\vspace{-0.2in}
\end{figure}

\noindent\textbf{Prior:} To model prior knowledge for the brain state classification problem, human expectations for statistical regularities in the environment and the uncertainty involved are summarized as a persistence diagram $D_X$. We assume that the underlying prior uncertainty of a persistence diagram $D_X$ is generated by a Poisson point process $\D_X$ with intensity $\lambda_{\D_X}$. An example of prior persistence diagrams is shown in Fig. \ref{fig:BayesTDA} (a) as black rectangles. 
Any point $x$ in a persistence diagram $D_X$ may not be observed in actual data due to the presence of noise, sparsity, and/or other unexpected scenarios. We address this instance by defining a probability function $\alpha(x)$. In particular, if $x$ is not observed in the data, the probability of this event is $(1-\alpha(x))$ and similarly $\alpha(x)$ is the probability of $x$ being observed. 

\noindent\textbf{Data/Likelihood Model:}
EEG signals are encoded into the observed PDs, $D_Y$, using the method discussed in Section \ref{sec:sublevel}. Points $y_i \in D_Y$  are linked to points in
PD $D_X$, generated by the prior PPP. We investigate the linking of these points to the prior PPP by relying on
the theory of marked Poisson point processes (MPPP) \cite{Kingman1993,Jacobson2005}. The probability density of the MPPP is given by a stochastic kernel, $\ell$ such that the marks $m(x_i)$ of $x_i$ are drawn independently
from  $\ell(\cdot|x_i)$, which in our case plays the exact role of the likelihood (see Section \ref{sec:ppp} for details). One needs to account for all possible marks, with the more likely marks realized as larger likelihood values $\ell(y_i|x_i)$ for all $(x_i,y_i) \in D_X \times D_Y$. In order to accommodate the nature of persistence diagrams, we need to define one last point process  that unveils the topological noise in the observed data. Intuitively, this point process consists of the points in the observed diagram that fail to associate with the prior. We define this as a Poisson point process $\D_{Y_U}$ with intensity $\lambda_{\D_{Y_U}}$.

%
A sample observed persistence diagram is shown in Fig. \ref{fig:BayesTDA} (a) as red hexagons. Fig. \ref{fig:BayesTDA} (b) and (c) show different combinations of possible associations between prior and data in the green regions. However, it is evident that the associations in (b) would have higher likelihood values than that in (c) and in turn, would have more impact on posteriors. Also, for every
configuration, some of the observed points do not associate with any point $x_i \in D_X$, which is shown with blue regions. We denote the features in blue regions as $D_{X_V}$, which stands for the features that vanished. If they are not vanished and make associations with features of $D_{Y}$, we denote it as $D_{X_O}$.    
Samples from  $\D_{Y_U}$ are shown in Fig. \ref{fig:BayesTDA} (b) and (c) as yellow regions.

\noindent\textbf{Posterior:}
With the above model characterization, the posterior intensity which explicitly show the update of the prior has the following form \cite{Maroulas2019a}: 
\begin{align} 
\small &\lambda_{\D_X|D_{Y_{1:m}}}(x) = \rhighlight{\left(1-\alpha(x)\right)\lambda_{\D_X}(x)} + \nonumber \\ &\frac{\alpha(x)}{m} \sum_{i=1}^m\!\!\sum_{y \in \D_{Y^i}}\!\!\!\frac{\bhighlight{\ell(y|x)\lambda_{\D_X}(x)}}{\brhighlight{\lambda_{\mathcal{D}_{Y_U}}(y)}+\bhighlight{\int_{\W}\ell(y|u)\alpha(u)\lambda_{\D_X}(u) du}}.\label{postrior_operator}
\end{align}
In the posterior intensity density, the two terms reflect the decomposition in the prior point process. 
The first term is for the features of prior which may not be observed and hence the intensity is weighted by $(1-\alpha(x))$. 
On the other hand, the second term corresponds to  the features in prior  that may be observed and  similarly is  weighted by $\alpha(x)$. Here we observe an expression consistent with the traditional Bayes' theorem, specifically a product of prior intensity and likelihood divided by a normalizing constant. The normalizing constant consists of two terms illustrating the two instances of our data model. $\D_{Y_U}$ consists of the features that are not associated to the prior and this is evident in the first term of the normalizing factor. Consequently, the second term provides the contribution of the observed  data from $\D_{Y}$, coupling with  prior features to form the marked PPP.

\section{Bayesian Classification \label{sec:class}}


In this section, we develop a Bayesian learning  approach that discriminates EEG signals from different cognitive states. 
In particular, we present a classification scheme based on Bayes factors of persistence diagrams generated from physiological signals. 
We start by extracting fundamental topological features from a collection of EEG signals and record the information in persistence diagrams using the sublevel set filtration discussed in Section \ref{sec:sublevel}. 

For a  persistence diagram $D$ that needs to be classified, we assume that $D$ is sampled from a Poisson point process $\mathcal{D}$ in $\mathcal{H}$ with prior intensity $\lambda_{\mathcal{D}}$. Consequently, its probability density has the form
$ p_{\mathcal{D}}(D)=\frac{e^{-\lambda}}{|D|!}\prod_{d \in D}\lambda_{\mathcal{D}}(d),
$
where $\lambda= \int_{\W} \lambda_{\mathcal{D}}(u)du$ is the expected number of points in $\mathcal{D}$. For training sets $Q_{Y^k} := D_{Y^k_{1:n}}$ for $k = 1, \cdots, K$ from $K$ classes of random diagrams $\mathcal{D}_{Y^k}$, we obtain the posterior intensities by following  the estimation process discussed in Section \ref{sec:main}. The posterior probability density of $\mathcal{D}$ given the training set $Q_{Y^k}$ defined as 

\vspace{-0.2in}
\begin{equation} \label{eqn:poisson_posterior_density}
p_{\mathcal{D}|\mathcal{D}_{Y^k}} (D|Q_{Y^k}) = \frac{e^{-\lambda}}{|D|!}\prod_{d \in D}\lambda_{D|Q_{Y^k}}(d).
\end{equation}

The posterior probability densities given the other training sets are obtained by analogous expressions. Consequently, the Bayes factor is defined as 

\vspace{-0.2in}
\begin{equation} \label{eqn:bayes factor}
BF^{i,j}(Q_{Y^i},Q_{Y^j})=\frac{p_{D|\mathcal{D}_{Y^i}}(D|Q_{Y^i})}{p_{D|\mathcal{D}_{Y^j}}(D|Q_{Y^j})}
\end{equation} 

For every pair of $(i,j)$ for $1\leq i,j\leq K$ if $BF^{i,j}(Q_{Y^i},Q_{Y^j})>c$, we will assign one vote to class $Q_{Y^i}$ or otherwise for $BF^{i,j}(Q_{Y^i},Q_{Y^j})<c$. The final assignment of the class of $D$ to a class is obtained by a majority voting scheme. 

%

\section{Application to EEG \label{sec:app}}

\subsection{ Conjugate family of priors for EEG signals \label{sec:gm_post}} 

%

Here, we present a a closed form of the posterior distribution through a conjugate family of priors, e.g., the Gaussian mixtures. Hence the prior-to-posterior updates yield posterior distributions of the same family.
We specify the prior intensity density as $\lambda_{\mathcal{D}_X}(x) = \sum_{j = 1}^{N}c^{\mathcal{D}_X}_{j}\mathcal{N}^{*}(x;\mu^{\mathcal{D}_X}_{j},\sigma^{\mathcal{D}_X}_{j}I)$,  where $N$  is the number of corresponding mixture components and $\mathcal{N}^{*}$ is the restricted Gaussian density on the wedge $\W$. In a similar fashion, we define the density of the Poisson point process $\mathcal{D}_{Y_U}$. The likelihood density is also Gaussian as $\ell(y|x) = \mathcal{N}^{*}(y;x,\sigma^{\mathcal{D}_{Y_O}}I)$  and $\alpha(x) =\alpha$. With all of these we obtain a Gaussian mixture posterior intensity density of the form

\vspace{-0.3in}
\begin{align}
\label{eqn:mg_posterior}
\lambda_{\mathcal{D}_X|D_{Y^{1:m}}}(x) & = (1-\alpha)\lambda_{\mathcal{D}_X}(x)+ \nonumber\\ &\frac{\alpha}{m}  \sum_{i=1}^m\!\!\sum_{y \in \D_{Y^i}}\!\!\sum_{j=1}^{N}\!\! C_{j}^{x|y}\mathcal{N}^*(x;\mu_{j}^{x|y},\sigma_{j}^{x|y}I),
\end{align}
where $C^{x|y}, \mu^{x|y}$ and $\sigma^{x|y}$ are weights, mean and variance of the posterior intensity respectively, corresponding to the second part of  \eqref{postrior_operator}, and these are pertinent updates of the prior parameters \cite{Maroulas2019a}.

\subsection{EEG Datasets \label{sec:data}}
US Army Aberdeen Proving Ground (APG) researchers have simulated noisy EEG signals based on different mental activities. We used this dataset for our analysis mainly focusing on two different frequency bands -- alpha and beta. Alpha (frequency from 8 to 13 Hz)  corresponds to intense mental activity, stress, and tension, and beta (frequency 13–30Hz) correlates with active attentions and focusing on concrete problems or solutions  \cite{Siuly2016}.
As the dataset contains several predominant oscillations based EEG signals, a Gaussian conjugate prior produces promising results for estimating the posterior probabilities as well as for Bayes factor classification \cite{Norwich1993,vanPutten2001,Ince2017}. 

\subsection{Posterior estimation of EEG Datasets \label{sec:eeg_post}} 
\begin{figure}[t]
	\centering
	\subfigure[]{\includegraphics[width=1in,height=0.8in]{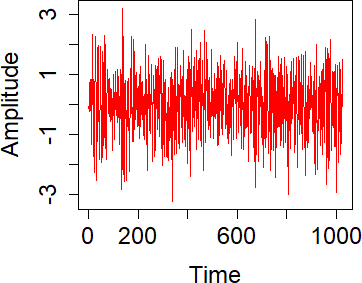}}\hspace{0.1in}
	\subfigure[]{\includegraphics[width=0.8in,height=0.8in]{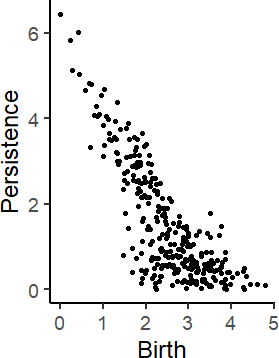}}\hspace{0.1in}
	\subfigure[]{\includegraphics[width=1in,height=1in]{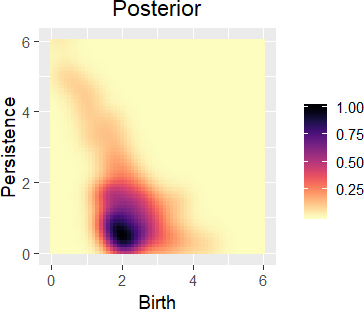}} 
	\subfigure[]{\includegraphics[width=1in,height=0.8in]{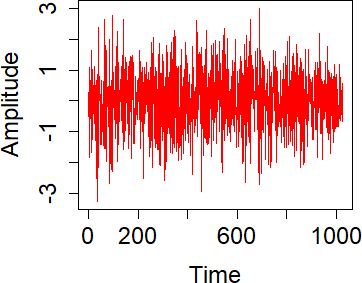}}\hspace{0.1in}
	\subfigure[]{\includegraphics[width=0.8in,height=0.8in]{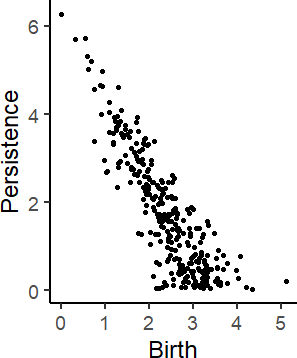}}\hspace{0.1in}
	\subfigure[]{\includegraphics[width=1in,height=1in]{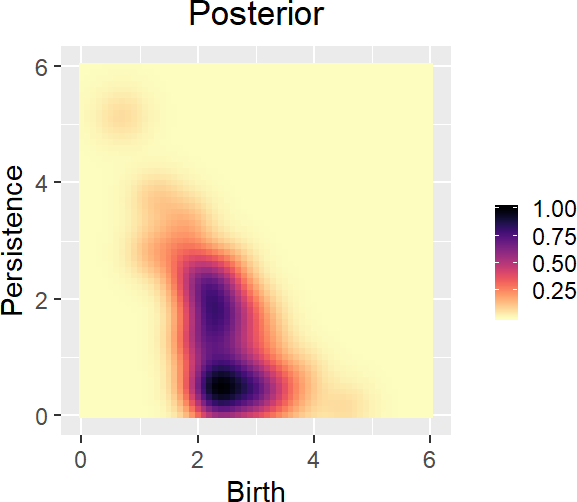}} 
	\caption{(a) is an alpha band simulated EEG signal, (b)  is the corresponding persistence diagram using sublevel sets and (c) is the posterior intensity map obtained from \eqref{eqn:mg_posterior}. Similarly (d) is a beta band, (e) is the  corresponding persistence diagram, and (f) is the corresponding posterior intensity map.}
	\label{fig:pd_signal}	
	\vspace{-0.2in}
\end{figure}

We first converted the EEG signals to persistence diagrams via sublevel set filtrations. In Fig. \ref{fig:pd_signal}, we present two samples from the EEG dataset of alpha (a) and beta (d) bands respectively along with their persistence diagrams in (b) and (e).
Typically EEG signals encode various forms of noise and the simulated EEG dataset accounts for this by corrupting these signals with additive noise. The signals in Fig. \ref{fig:pd_signal} have the signal to noise ratio (SNR) 0, which implies equal contribution from signal and noise. 


In Fig. \ref{fig:pd_signal} we illustrate a posterior intensity estimation  of a noisy alpha band  and a noisy beta band  utilizing \eqref{eqn:mg_posterior}. 
To demonstrate a data-driven posterior, we employed an uninformative prior of the form $\mathcal{N}((3,3),20I)$. 
To present the intensity maps uniformly, we divide the intensities by their corresponding maxima and call them scaled intensities ranging from 0 to 1.

\subsection{EEG signal classification with Bayesian learning \label{subsec:example_class}}
Detection and classification of specific patterns in the brain activity are crucial steps in understanding functional behaviors for developing human-machine communications. We have taken the first step toward engaging Bayesian learning in EEG signal analysis by implementing Gaussian posterior intensities as explained in Section \ref{sec:gm_post} and using these posteriors for a binary Bayes factor classification.
From the dataset provided by APG researchers, we used two instances of additive noise in order to represent cases with two different SNR. Our considered dataset comprises SNRs such as 3 and 5, where SNR 5 has more contribution from the signal than SNR3.

We followed the process discussed in Section \ref{sec:gm_post}  to estimate the posterior intensity of a persistence diagram $D$  in $\mathcal{H}$ given a training set $Q_Y$, with the goal of identifying the correct class of $D$. We used the R package \href{https://github.com/maroulaslab/BayesTDA}{BayesTDA}
to obtain posterior intensities. Consequently, the probability density was obtained from \eqref{eqn:poisson_posterior_density}. 
After computing the intensities with respect to the training sets from both of the classes, the Bayes factor was computed by \eqref{eqn:bayes factor} as the ratio of the posterior probability densities of the unknown persistence diagram $D$ given each of the two competing training sets from $Q_{Y}$ or $Q_{Y'}$.  For a threshold $c$, $BF(D)>c$ implies that $D$ belongs to $Q_{Y}$ and $BF(D)<c$ implies otherwise. 

\begin{figure}[t]
	\centering{
		\includegraphics[width=2.5in,height=2in]{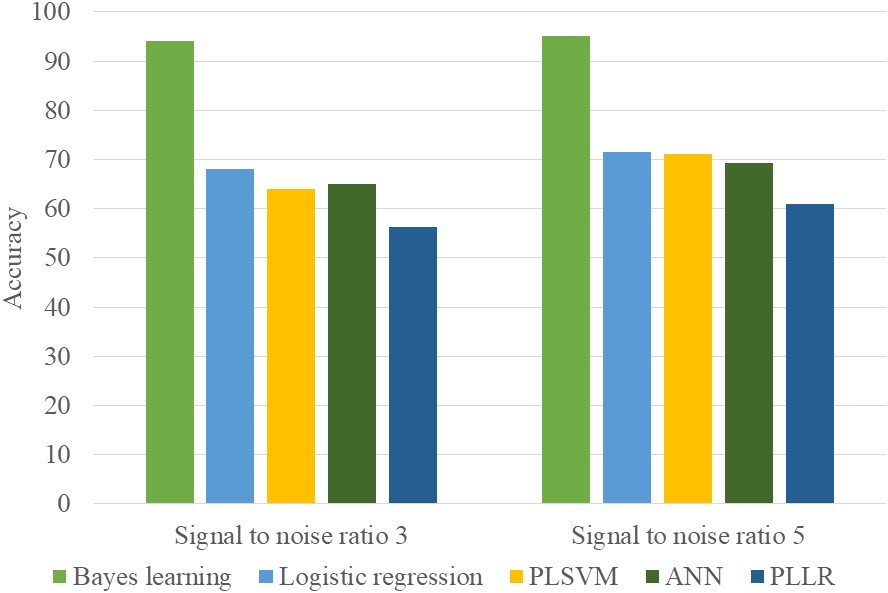}
	}
	\caption{Comparison of our Bayesian learning method with  logistic regression, Artificial Neural Network (ANN), persistence landscape with support vector machine (PLSVM), and  persistence landscape with logistic regression (PLLR). ANN was trained by a standard back propagation algorithm, a sigmoid activation function, and some other parameters such as, number of hidden layers = 100, the maximum number of iterations =1000, error threshold =0.001 and learning rate =0.1.\label{fig:comp} }
		\vspace{-15pt}
\end{figure}

We implemented 10-fold cross validation for estimating the accuracy.
For this we partitioned each class
into 10 different sets and 9 of them for each class were used for training
sets, and 1 was used for testing. We repeated this 10 times so that every partition  acts as the testing data exactly once. We then found the average among all partitions. Results from the Gaussian learning scheme are presented in Fig. \ref{fig:comp}. 
We compared the results of Gaussian learning scheme with Artificial Neural Networks (ANNs), logistic regression (LR) with features (mean, standard deviation and entropy of the recorded coefficients) extracted from Wavelet Transform (WT). We prefer to use WT rather than Fourier transform (FT) due to its inability to analyze nonstationary nature of EEG signals \cite{Al-Fahoum2014,Fiscon2018}. Both ANN and LR have been widely applied for physiological signals classification \cite{Dindin2019,Sivasankari2014,Subasi2005,Tomioka2007,Kabir2016,Prasad2014,Bahy2016}. We also compared our result with an exiting TDA technique namely, persistence landscape \cite{Bubenik2015}.  We extracted the first landscape functions of the persistence diagrams for all considered EEG signals and implemented support vector machine and logistic regression  on the extracted landscape function. 
 Our results for classifying these two bands of SNRs  outperforms the other existing TDA and non-TDA based classification methods over all levels of SNRs considered here. Furthermore, the Gaussian learning scheme is able to classify almost perfectly with a high signal to noise ratio.

\section{Conclusion \label{sec: conclusion}}
In this work, we have proposed a  Bayesian framework for persistence diagrams that incorporates prior beliefs about signals and does not rely on any regularity assumptions such as stationarity or linearity for the computation of posterior distributions. The topological descriptors, e.g., persistence diagrams  
of EEG signals can decipher essential shape peculiarities by avoiding complex and unwanted geometric features.  
Our method perceives persistence diagrams as point processes (PPs).
As required for a Bayesian paradigm, we incorporate prior uncertainty by viewing persistence diagrams as Poisson PPs with a given intensity. 
We model the connection between prior PP and persistence diagrams of noisy observations through marked PPs. These models the data likelihood component of the Bayesian framework. Additionally, we define the likelihood through topological summaries of a signal rather than using the entire signal. This is analogous to the substitution likelihood discussed by Jeffreys \cite{Jeffreys1961}. 

Relying on the posterior distributions obtained from the Bayesian framework, we develop a Bayesian learning scheme. Furthermore, we present a closed form of the posterior estimation through a conjugate family for priors. In the case of synchronized brain activity, this implementation is useful for analyzing EEG signals. This exhibits the ability of our method to recover the underlying persistence diagram, analogously to the standard Bayesian paradigm for random variables.

We employ this  Bayesian learning scheme for EEG signal classification. 
We provide a detailed comparison with some of the existing methods of signal classification and showcase that our method outperforms them. For comparison purposes, we pursue two directions. Firstly, we compare with two most widely used signal classification algorithms--neural nets and logistic regression. Secondly, we show a comparison between our method and another topological tool, namely persistence landscape, with traditional machine learning methods such as support vector machine and logistic regression. We exhibit higher accuracy for all considered cases.     
%
Thus, the Bayesian inference developed here opens up new avenues for machine learning algorithms for complex signal analysis \emph{directly} on the space of persistence diagrams. 

\bibliographystyle{Ieeetran}
\bibliography{bib1}


\end{document}